\documentclass[10pt,twocolumn,letterpaper]{article}

\usepackage{wacv}
\usepackage{times}
\usepackage{epsfig}
\usepackage{graphicx}
\usepackage{amsmath}
\usepackage{amssymb}
\usepackage{siunitx}
\usepackage{subfig}
\usepackage[numbers,sort&compress]{natbib}
\usepackage[usenames, dvipsnames]{color}
\usepackage{pifont}
\newcommand{\cmark}{\color{ForestGreen}\ding{51}}%
\newcommand{\xmark}{\color{red}\ding{55}}%

\newcommand{\rulesep}{\unskip\ \vrule\ }

\wacvfinalcopy 

\def\wacvPaperID{448} 

\ifwacvfinal\fi
\begin{document}

\title{On measuring the iconicity of a face}

\author{Prithviraj Dhar \hspace{2cm} Carlos D. Castillo \hspace{2cm} Rama Chellappa\\
University of Maryland, College Park\\
{\tt\small \{prithvi,carlos,rama\}@umiacs.umd.edu}
}

\maketitle
\ifwacvfinal\thispagestyle{empty}\fi

\begin{abstract}
For a given identity in a face dataset, there are certain iconic images which are more representative of the subject than others. In this paper, we explore the problem of computing the iconicity of a face. The premise of the proposed approach is as follows: For an identity containing a mixture of iconic and non iconic images, if a given face cannot be successfully matched with any other face of the same identity, then the iconicity of the face image is low. Using this information, we train a Siamese Multi-Layer Perceptron network, such that each of its twins predict iconicity scores of the image feature pair, fed in as input. We observe the variation of the obtained scores with respect to covariates such as blur, yaw, pitch, roll and occlusion to demonstrate that they effectively predict the quality of the image and compare it with other existing metrics. Furthermore, we use these scores to weight features for template-based face verification and compare it with media averaging of features.
\end{abstract}

\vspace{-4mm}
\section{Introduction}
\begin{figure}
{\centering
\subfloat[]{\includegraphics[width = 0.30\linewidth, height = 20mm]{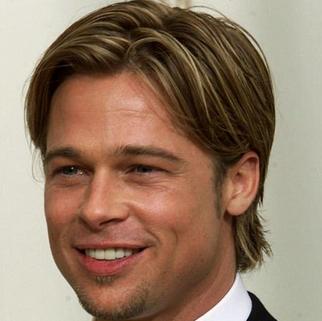}}
\rulesep
\subfloat[]{\includegraphics[width = 0.30\linewidth, height = 20mm]{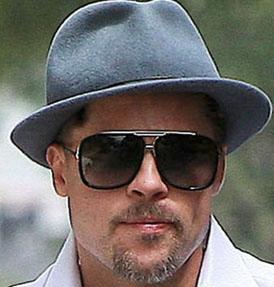}}
\caption{(a.) Iconic and (b.) Non-iconic image of Brad Pitt. Our approach assigns an iconicity score of 0.84 and 0.33 to (a) and (b), respectively.}
\label{fig:bpitt}
}
\end{figure}

What makes Brad Pitt look like Brad Pitt? A clean frontal image of the actor might better represent him, than an image where he is wearing sunglasses and a large fedora (Figure \ref{fig:bpitt}). In this case, the former can be considered as an iconic image and the latter as a non-iconic one. Predicting face iconicity is a useful task in facial image analysis. This problem is difficult because `iconicity' is subjective, and is dependent on the existing images of a subject. Most face recognition and verification systems are known to perform well for iconic images captured in constrained environments. However, for measuring the performance of such systems in real life scenarios, they should be evaluated on unconstrained faces. Moreover, the performance of a system can be accurately measured by taking into consideration the difficulty (based on the iconicity) of the test dataset. So, computation of iconicity of a given facial image is useful for properly evaluating face verification and recognition systems.\\
In \cite{berg2007automatic}, an iconic image for an object is defined as an image with a large clearly delineated instance of the object in a characteristic view. Here the authors showed that iconic images can be identified rather accurately in natural datasets by segmenting images with a procedure that identifies foreground pixels. But this does not translate well to identifying iconic facial images. One unsupervised method to segregate iconic and non-iconic face images would be to perform global clustering across all identities and conclude that the images that are not present in the appropriate identity cluster are non-iconic.  However, in this setting, we cannot compute iconicity of any unseen identity. Moreover, in a dataset containing millions of images (such as UMD Faces, MS-Celeb-1M) it can be very cumbersome to perform clustering. Thus it would be very helpful to have a method which can assign an iconicity measure on any unseen image without any additional information. \\

It can be expected that an iconic image is likely to be a high quality face image. Thus, computing iconicity can also help us develop a notion of face quality. In \cite{best2017automatic}, the authors propose a framework to regress the quality scores of a model to human quality values. Although this method can directly estimate the face quality of unseen images, its training phase is not scalable, as it is expensive to assign humans to provide ground truth for a regression model. Hence, it is important to design an approach where a blind quality/iconicity prediction model can be trained only by optimizing an objective that is only dependent on the inherent properties of a given image. \\

Similarity of an image with another is one such inherent property. In this work, we define iconicity as the verifiability of an image and propose a technique to use pairwise similarity for training a model to predict the verifiability of an image. This model is built as a Siamese Multi Layer Perceptron (MLP) network, which is trained using deep face descriptors extracted using a recognition network and optimized using their similarity. It should be underscored that `verifiability' of an image (also known as its iconicity) is defined as the distinct suitability of a facial image feature to be matched with any other image feature of the same identity, in the entire dataset. The training of Siamese MLP does not require any iconicity-based supervision, and only needs identity labels. This is important as there exist several face datasets with identity labels but no datasets with iconicity-based annotation.\\

During training, the model learns to map a given feature to iconicity scores by learning its interaction with those of same and different identities. Once the model learns this mapping, it can be used to predict the iconicity of images of unseen identities. After the model is trained, the measure of iconicity is unary. At test time, we do not require any explicit information such as identity labels, the global knowledge of the dataset or any predefined reference images and attributes. We can also generate different sets of iconicity scores corresponding to different sets of descriptors for training the Siamese MLP network.

The contributions of our work are as follows: (1) We propose a novel method to relate the iconicity of a face image to verifiability of its descriptor. (2) We propose a simple Siamese Multi Layer Perceptron architecture to estimate the iconicity of a given facial image feature, without any iconicity-based supervision. (3) We demonstrate that defining iconicity in the aforementioned manner correlates well with factors that affect the visual quality of an image. (4) We establish a use-case for our iconicity scores by using them to pool features in template-based face verification, and obtain results comparable to state-of-the-art.   

\begin{figure}
    \includegraphics[width=\linewidth,height=25mm]{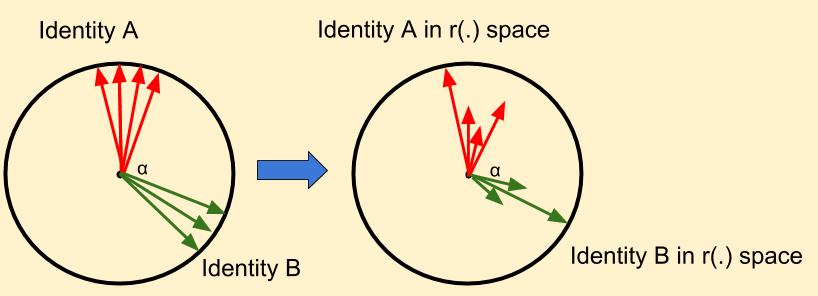}\vspace{-2mm}
    \caption{We propose an approach to map feature vector of a facial image whose length is a function of the iconicity}
    \label{fig:sphere}
\end{figure}

The paper has been organized as follows: Previous work in this area is summarized in Section 2. We present our approach in Section 3 and describe the experiments performed in Section 4. Experimental results are reported in Section 5.

\section{Related work}
\label{sec:prev}
Face iconicity (or image iconicity in general) has not been explored widely in previous research. In \cite{berg2009finding}, the authors use compositionality of an image, and its similarity with images of different object categories to select images that are highly representative of the class. This is followed by clustering to filter out the most iconic images of an object category. But, the specified method requires explicit information about categories at test time.

As mentioned earlier, face iconicity is related to face quality. We believe that an ideal face iconicity metric should correlate well existing image quality metrics. Before we proceed, it is important to highlight that image quality assessment (IQA) and face image quality prediction are different tasks. For instance, a profile face image can be of high quality, yet it might not represent the identity as well as a frontal image can. Also, it should be noted that IQA is usually defined in the context of image compression. However, due to the shared attributes in face quality prediction and general IQA, we discuss the relevant work in both of these areas.
Existing literature on (facial) image quality can be divided into three categories which are discussed below: \\ 
\textbf{Techniques computing a specific attribute to measure quality} : Previous experiments (\cite{abaza2012quality}, \cite{wang2004image}, \cite{sellahewa2010image}) in this field have defined quality on the basis of a specific attribute of the facial image, such as pose, blur etc. For example, the authors of \cite{abaza2012quality} defined quality on the basis of contrast, brightness, focus, illumination and sharpness.  The authors in \cite{mittal2012no} propose the BRISQUE algorithm based on the `naturalness' of the image, defined as the deviation of the normalized luminance distribution from its Gaussian estimation. However, these approaches use a specific attribute or reference images to develop a notion of face quality, and hence cannot be scaled to larger datasets with more variation and breadth. 

\textbf{Techniques which utilize explicit manually extracted quality specific information:} Several experiments and most of the CNN/deep network-based quality assessment frameworks require supervision with respect to quality score during training. However, the size of such datasets is limited and construction of such databases is time-consuming and expensive. Moreover, there does not exist such datasets for face quality. For example, in \cite{kang2014convolutional}, the authors proposed a no-reference deep CNN architecture which is trained on datasets (such as LIVE \cite{sheikh2005live}), that supply image degradation level as ground truths. Similarly, the authors proposed a learning-to-rank technique in \cite{chen2015face}, where in weights (which indicate quality) are learned for image features according to their respective datasets, such that the predefined ranking of the quality of images in the datasets considered is maintained. However, the ranking of the dataset quality needs to be explicit in this work. In \cite{liu2017rankiqa}, the authors used a Siamese network to rank the image features according to their quality. However, their approach explicitly required quality-based ranks for training. Moreover, as their framework is not specifically defined for faces, the feature similarity cannot indicate of the pair-quality as there is no concept of identities.\\ 

\vspace{-1.95mm}
\textbf{Descriptor/image-to-score techniques} One of the more recent works \cite{ranjan2018crystal}, used the scores assigned to a face by a face detector (FD scores) as a measure of facial image quality. In this work it has been established that weighting features of network proposed in \cite{ranjan2017l2} with these quality scores helps to improve the performance of existing template-based face verification systems. Another interesting work in this area is \cite{parde2017face}, where it was proposed that the norm of a facial image feature can be used as a quality measure of the face. It was demonstrated that a lower norm value is associated with low quality images. However, this is not applicable for normalized image features (such as those obtained in \cite{ranjan2017l2}, \cite{wang2018cosface}, \cite{wang2017normface} etc. where the network normalizes the feature as part of the training process). To the best of our knowledge, \cite{parde2017face} and \cite{ranjan2018crystal} are the only works for face quality prediction (as opposed to general IQA), where computation of the quality scores is completely independent of reference images/attributes and/or does not rely on quality level ground truth during training. Hence, we present a technique in this category which directly predicts the iconicity of an unseen facial image given its feature descriptor. This alleviates the dependence of the predictor on reference attribute or image, making the method scalable and generalizable.
\section{Our Approach}
\begin{figure}
    \includegraphics[width=\linewidth, height=50mm]{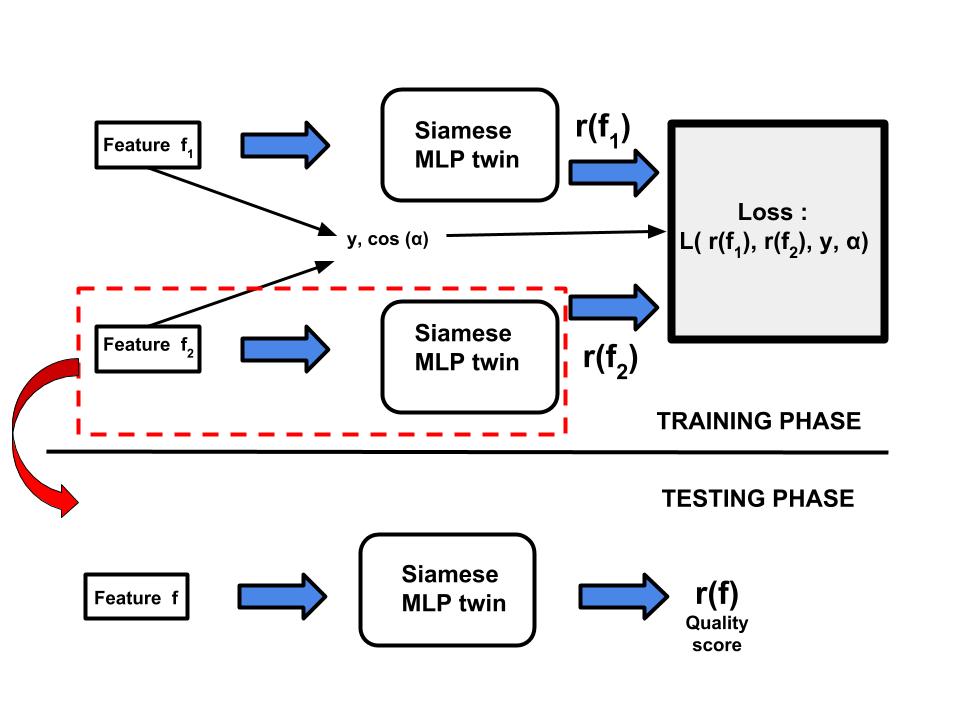}\vspace{-6mm}
    \caption{Overview of our proposed approach. During training, we compute the loss using cosine similarity $\cos\alpha$, pair label $y$, which guides the growth/decay of $r(f_1)$ \&  $r(f_2)$. $f_1$ and $f_2$ are extracted from a recognition network. During testing, we use one of the trained twins. }
    \label{fig:arch}
\end{figure}
We view iconicity as one of the indicating factors of the quality of an image. We hypothesize that for an identity consisting a mixture of iconic and non-iconic images, if a facial image cannot be matched successfully with images of the same identity, then it is a non-iconic image, i.e. verifiability of that image is low. This information from the training dataset is indirectly used to optimize an objective function, which is designed to enable the network estimate the verifiability of a feature.  Once the model learns to map a feature to its verifiability, it can predict the same for any feature, irrespective of identity, during evaluation. We use deep feature representations of faces instead of the raw image and propose an approach (depicted in Figure \ref{fig:arch}) to predict the verifiability of any given image feature, without using reference images/attributes.
\begin{table*}[t]
\centering
\begin{tabular}{ ||c c c c|| }
 \hline
 \multicolumn{4}{|c|}{Interpretation table} \\
 \hline
 Type& $\mathbf{\cos(\alpha)}$ & $\mathbf{y(\Delta -r(f_1)r(f_2).\cos\alpha)}$ & Effect on $\mathbf{r(f_1)r(f_2)}$ for optimization\\
 \hline
 I   & Relatively low ($<0$- Non iconic)    &Relatively high&   Decrement\\
 II (Disguise)   & Relatively high ($>0$- Non iconic)    &Relatively high&   Decrement\\
 III&   Relatively high ($>0$- Iconic)  & Relatively low   & Increment \\
 IV& Relatively low ($<0$- Iconic) & Relatively low& Increment  \\
 \hline
\end{tabular}
\vspace{-2mm}
\caption{Tentative effect of $\cos \alpha $ on the product ($r(f_1) \, r(f_2)$) }
\label{table:intpret}
\end{table*}
\subsection{Estimating iconicity using pairwise learning}
\label{sec:pairwise}
On a hypersphere of images, the cosine similarity of an image feature pair provides an estimate of the angular separation between these images. However, in this hypersphere, the notion of length of feature vectors is lost as the cosine similarity is obtained by normalizing the feature vectors followed by their inner product. We present an approach which maps this representation to another hypersphere (shown in Figure \ref{fig:sphere}) where the length of the feature vector $f$ represents the iconicity of the feature and is represented by a function $r(f)$, which is the tentative output of our model. For this mapping we need to normalize the feature $\frac{f}{\|f\|}$ and represent it as $r(f).\frac{f}{\|f\|}$ in the new hyperspace, where $r(f)$ represents the iconicity of the feature.  Thus, the iconicity determines the length of the feature from the origin of the hypersphere. In order to learn $r(f)$, we use a pairwise learning technique where we optimize $\langle f_1, f_2 \rangle_r$, which represents the dot product of a feature pair in the new hypersphere ($r(\cdot)$) space. $$\langle f_1, f_2 \rangle _r = \bigg(r(f_1)\frac{f_1}{\|f_1\|}\bigg)^T\bigg(r(f_2)\frac{f_2}{\|f_2\|}\bigg)$$ $$\implies \langle f_1, f_2 \rangle _r = r(f_1) \, r(f_2)\, \cos\alpha$$ To optimize this inner product in the new space we formulate it as a new similarity measure which can take into account the iconicity of individual features as well. So, we define an objective function for training the model as:
\begin{equation}
\label{eq:loss}
L(f_1,f_2) = \max(0,\,\, y\,(\Delta -r(f_1)\, r(f_2)\, \cos\alpha))
\end{equation}
where $\Delta (> 0)$ represents the margin, which is a hyperparameter. Specifically, we want the angle $\alpha$ between two image features $f_1$ and $f_2$ and the label $y$ associated with the pair (+1 for positive pair i.e. both images from the same identity and -1 for negative pair i.e. if both images belong to different identity), to guide the growth/decay of $r(f_1)$ and $r(f_2)$. We build a Siamese MLP network to learn the function $r(\cdot)$. While training, it takes in a pair of features $f_1$ and $f_2$, and predicts their iconicity i.e. $r(f_1)$ and $r(f_2)$. Initially, $r(f)$ is a random scalar but as the network is trained, for a given feature $f$, $r(f)$ is optimized in a way such that it captures the verifiability of $f$ with respect to the entire dataset, by taking into consideration its associated pair label and its interaction with other features. Hence the model is trained to predict the verifiability of any given $f$, using angular separation and pair labels. The numerical interpretation of our loss function is provided in the next section. 

\subsection{Interpretation}
\label{sec:inter}
Without loss of generality, we can assume the presence of the following four types of pairs in the training dataset:
\textbf{Type-I} : At least one unclean image in a positive pair. Similarity scores of such pairs are less than zero, even though the associated images are of the same identity. \\
\textbf{Type-II }: At least one unclean image in a negative pair. Similarity scores of such pairs are positive, even though the associated images belong to different identities, representing disguise. \\
\textbf{Type-III }: Both clean images in a positive pair. Similarity scores of such pairs are positive, as expected.\\ 
\textbf{Type-IV }: Both clean images in a negative pair. Similarity scores of such pairs are negative, as expected. \\
Table \ref{table:intpret} illustrates the effect of $\alpha$ and the possible effect of loss function on the product $r(f_1) \, r(f_2)$. It can be noticed that the model is inclined to decrease this product if the pair contains at least one non iconic image.  
\subsection{Decoupling due to pairwise training}
\label{sec:dec}
In the previous sub section, we inferred that the product $r(f_1)r(f_2)$ would be decreased by the model if the pair contains at least one non-iconic pair. If the training dataset consists of $l$ iconic and $m$ non iconic images, then a given non iconic image feature $f$ can be associated with at most $l+m-1$ pairs, each of them belonging to either Type I or Type II. Similarly, a given iconic image can be associated with only $m$ pairs belonging to Type I or II. Hence during training, the product $r(f_1)r(f_2)$ involving a given non iconic image would be penalized more than that involving a given iconic image since $m \ll l+m-1$ i.e. the dataset consists of a mixture of iconic and non-iconic images. From this, we cannot directly deduce that the score of the non-iconic feature $f$ would be penalized more. This is because the product can also be decreased by penalizing the score of an iconic image and increasing that of the non iconic image in such a way that the product is decreased. However, as presented in Table \ref{table:intpret}, the product should be maximized when an iconic pair is encountered during training, to decrease the loss. So, for penalizing a product consisting of a non iconic image feature, the score of the non iconic image feature would be needed to be decreased. Therefore, it can be concluded that since a given non iconic image can be associated with more non iconic pairs (as  compared to iconic image), it would be penalized more than a given iconic image. 
\vspace{-3.8mm}
\subsection{Mixture of iconic and non-iconic images}
\label{sec:mix}
As explained in the previous subsection, in order to optimize the network, the number of iconic images $l$ and non iconic images $m$ should be chosen such that $m << l+m-1$, i.e. $l >> 1$. When sampling positive pairs, this implies that for a given identity, there should be several (more than one) iconic images. Similarly, when sampling negative pairs the network must encounter several iconic images. We empirically verified that choosing identities with $l$ and $m$ satisfying $\frac{m}{l+m} \approx 0.5$ during training yields the best results. This explains the requirement of mixture of iconic and non-iconic pairs for every identity. This step is performed before computing facial iconicity. Hence, we use Face Detection scores for calculating $l$ and $m$.
\vspace{-1.5mm}
\section{Experiments}
\subsection{Datasets}
As explained in Section \ref{sec:mix}, we require a training dataset that has a mixture of iconic and non iconic images for every identity. Hence we use Batch 1 of UMD Faces \cite{bansal2017umdfaces} which satisfies this requirement. This subset of the dataset consists of 175,534 images for 3674 subjects. For testing, we choose datasets with domains different from that of UMD faces. Since we show the correlation of our iconicity score with respect to several parameters (like blur, pose etc.), we select datasets with appropriate annotation of these parameters.

\subsection{Architecture - Siamese MLP}
\begin{figure}
    \includegraphics[width=\linewidth, height=40mm]{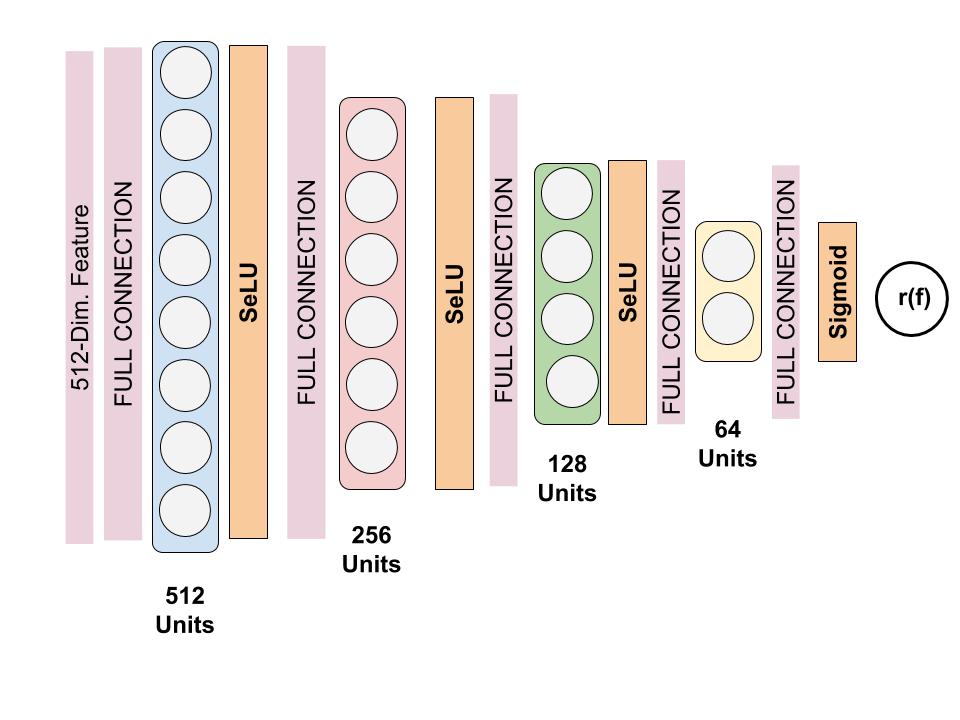}\vspace{-3mm}
    \caption{One of the Siamese MLP twins}
    \label{fig:netarch}
\end{figure}

We train a Siamese MLP network (Figure \ref{fig:netarch}), using pairs from Batch 1 of UMD Faces dataset. A single twin of the Siamese network accepts a 512 dimensional descriptor, and has 4 hidden layers, each consisting of 512, 256, 128, 64 hidden units. The first three hidden layers are followed by SeLU \cite{klambauer2017self} activations. There exists full connection between outputs these activation layers and the consequent hidden layer. The final hidden layer is then connected to the output node, which is scaled between 0 to 1 by using sigmoid unit. We assign a ground truth $y \in $ \{+1,-1\} on positive and negative pairs. Hence, Eq. \ref{eq:loss} is used as loss function. In our experiments, we chose $\Delta = 0.5$ based on the distribution of the similarity scores of the training dataset.

\subsection{Features used}
In order to demonstrate the robustness of our model across  different set of features, we perform experiments using features extracted using two separate network architectures. We use features of the architecture proposed in \cite{ranjan2017l2}, which is trained using $L_2$ constrained softmax loss, on MS Celeb 1M dataset \cite{guo2016ms}. All features generated using this network have unit norm. We also perform experiments with features learned by the network proposed in \cite{sankaranarayanan2016triplet}, which follows the AlexNet architecture. This network is trained on the CASIA-Webface dataset \cite{yi2014learning}. Thus, we train the following two Siamese MLP models, using two set of feature descriptors:\\
a.) \textbf{Model-1} : Siamese MLP trained with features of \cite{ranjan2017l2}\\
b.) \textbf{Model-2} : Siamese MLP trained with features of \cite{sankaranarayanan2016triplet}.\\
The aforementioned features are used as their corresponding networks have demonstrated high performance for the task of face verification. Also, in all our experiments, features are extracted after computing face coordinates using the all-in-one network described in \cite{ranjan2017all}.

\subsection{Training and testing}
For every epoch, we randomly choose 20000 positive and 20000 negative pairs from the dataset, feed the feature pair to the network and perform mini batch gradient descent using a batch size of 256. We train the network for 50 epochs. We train two models, using features of the network in \cite{ranjan2017l2} and \cite{sankaranarayanan2016triplet}. For testing we only use one of the twins of the trained model to determine the iconicity of any given image feature. 
\vspace{-4mm}
\section{Results}
\begin{figure*}
\centering
\subfloat[]{\includegraphics[width = 0.24\linewidth, height=30mm]{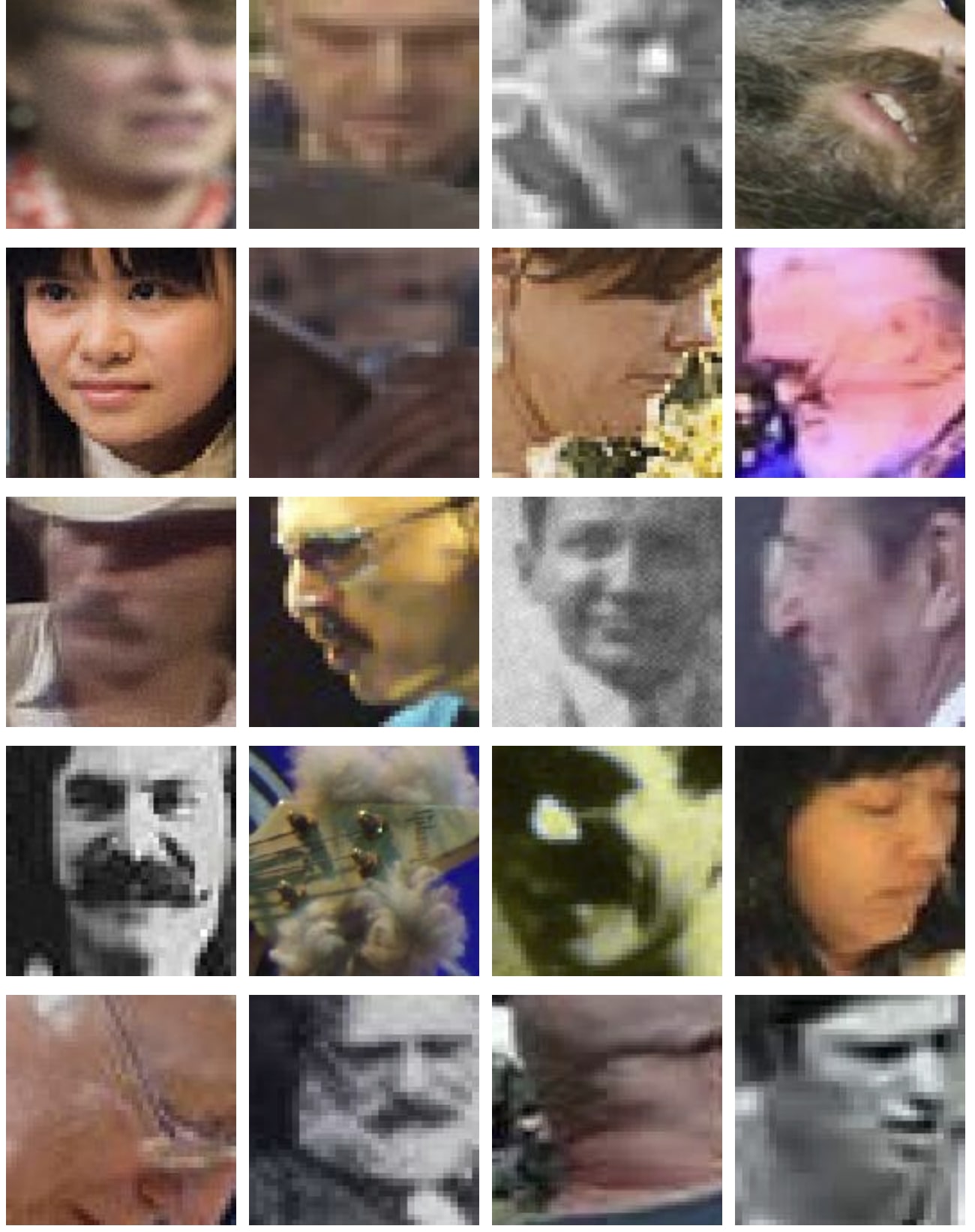}}
\rulesep
\subfloat[]{\includegraphics[width = 0.24\linewidth, height=30mm]{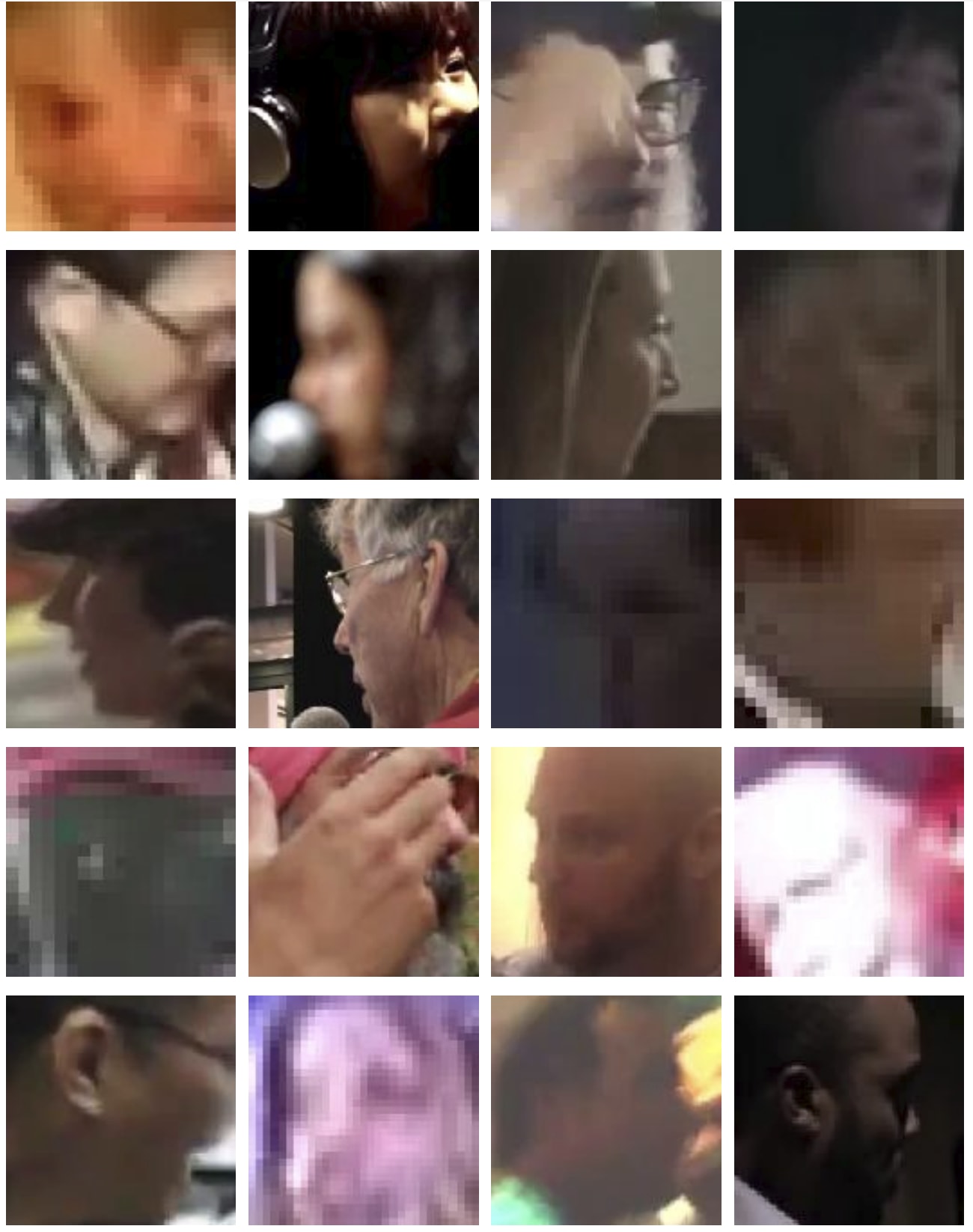}}
\rulesep
\subfloat[]{\includegraphics[width = 0.24\linewidth, height=30mm]{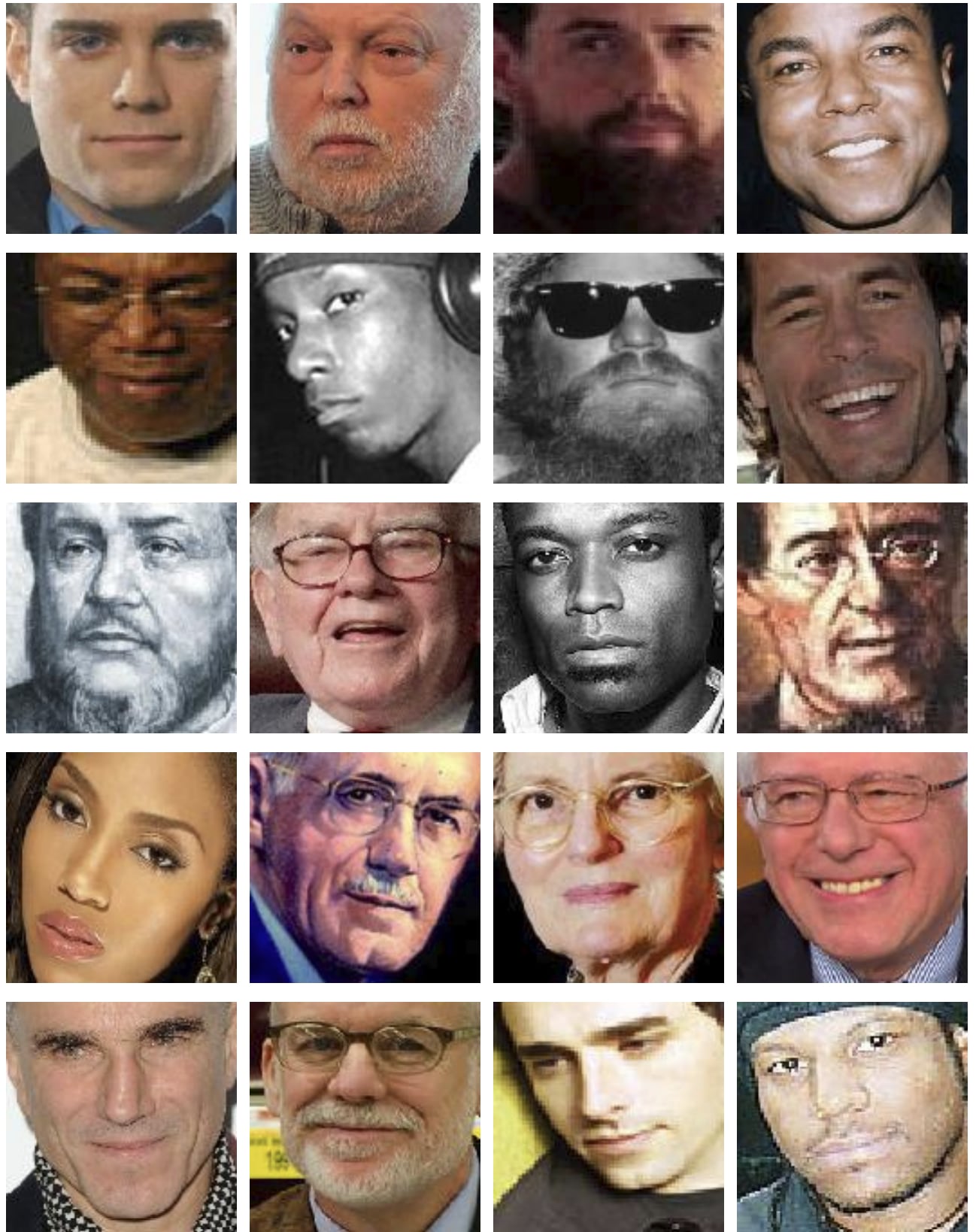}}
\rulesep
\subfloat[]{\includegraphics[width = 0.24\linewidth, height=30mm]{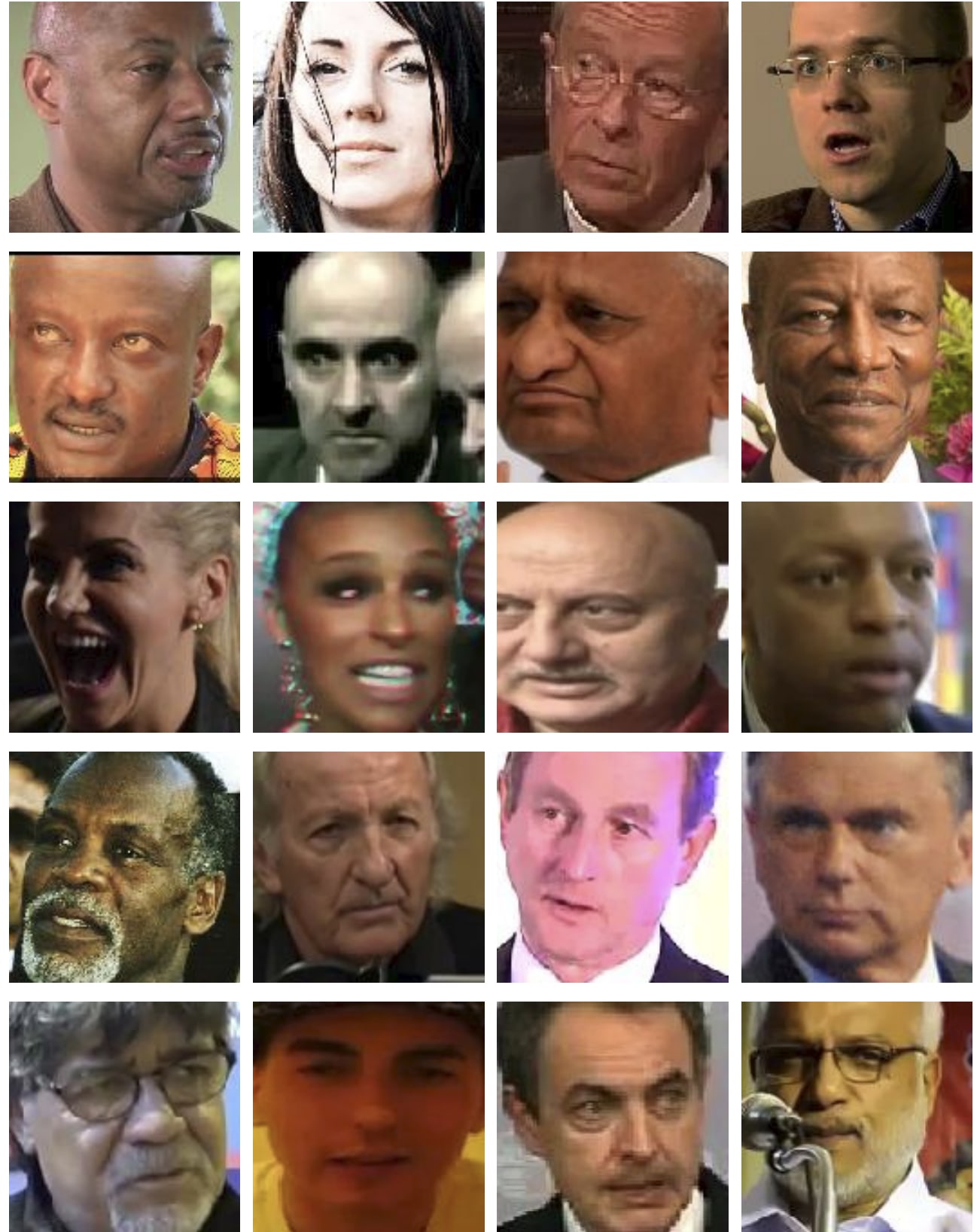}}\vspace{-4mm}
\caption{Images with low Model-1 iconicity score (i.e. $r(f)$) in a.) Batch 2 of UMD Faces \cite{bansal2017umdfaces}, b.) IJB-C dataset \cite{maze2018iarpa} and images with high Model-1 iconicity score in c.) Batch 2 of UMD Faces, d.) IJB-C dataset.}
\label{fig:visual}
\end{figure*}
\begin{figure*}
\subfloat[]{\includegraphics[width = 0.33\linewidth, height=35mm]{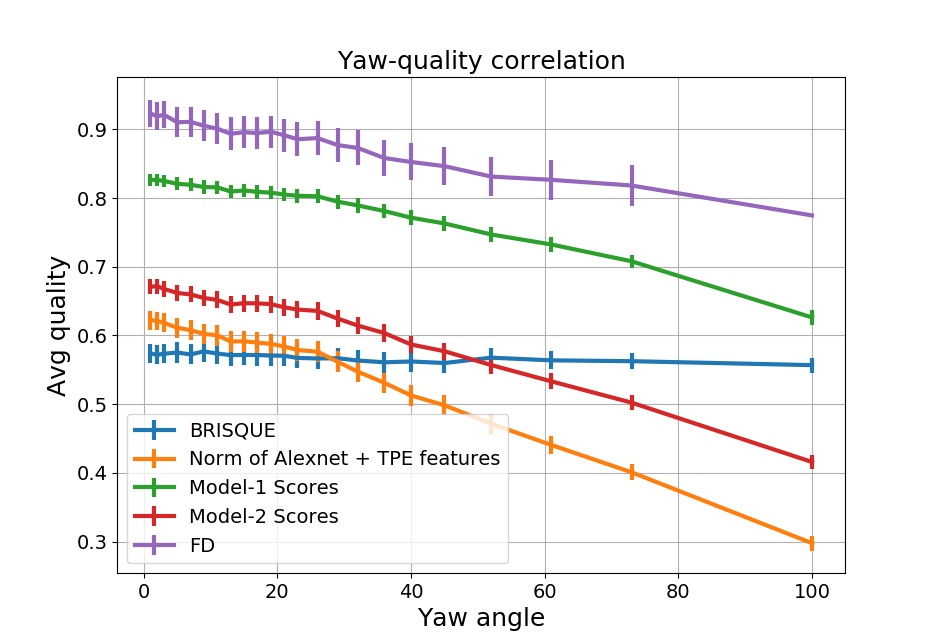}}
\subfloat[]{\includegraphics[width = 0.33\linewidth, height=35mm]{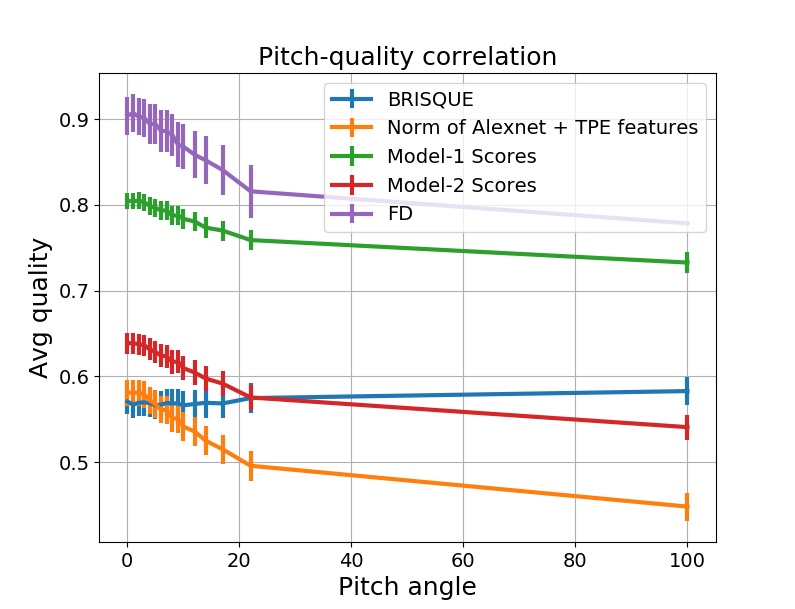}}
\subfloat[]{\includegraphics[width = 0.33\linewidth, height=35mm]{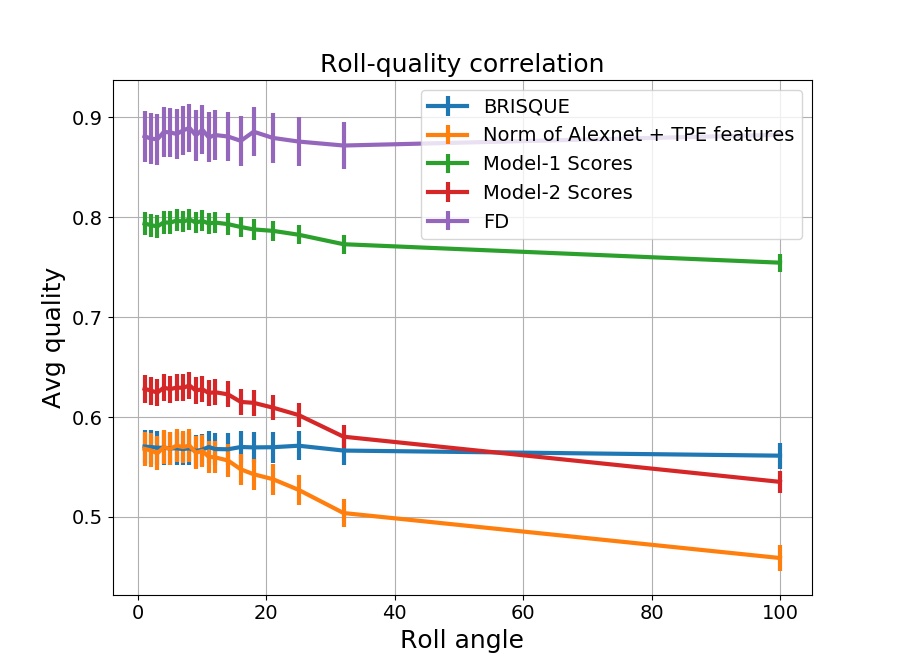}}\vspace{-4mm}
\caption{Variation of scores obtained with Model-1, Model-2, FD Score \cite{ranjan2018crystal}, norm of \cite{sankaranarayanan2016triplet} and BRISQUE \cite{mittal2012no} across  a.) Yaw, b.) Pitch, c.) Roll, on IJB-C dataset.}
\label{fig:posecorr}
\end{figure*}
An ideal face quality/iconicity metric is expected to be an all-in-one metric which obtains considerable performance on a variety of quality related tasks, and correlates well with human quality opinions. We evaluate face quality metrics on the basis of their performance on the following two tasks.\\
\textbf{Relation with factors that affect face quality:} To verify the usefulness of any quality score, we observe its correlation with some of the factors that visibly affect facial image quality : yaw, pitch, roll, occlusion and blur. We also demonstrate that our Siamese models generate iconicity scores well correlated with human perception of quality.

\textbf{Template-based face verification}: In \cite{best2017automatic} and \cite{ranjan2018crystal}, template-based face verification has been used as one of the applications of face quality measure. Hence, we also evaluate the effectiveness of our iconicity scores for this task. We use iconicity scores as weights to perform quality pooling for template-based face verification, and compare our verification results with that obtained with media averaging technique to combine features of the same template.

As mentioned in \cite{phillips2009introduction}, computing quality using a biometric system is a Biometric complete problem, which basically implies that our computed iconicity score cannot improve the one-to-one verification algorithm. However, in template-based face verification, the objective is to reduce the error rate of the system (and not to remove the weakness of algorithm). Thus, we expect our scores to help us appropriately weight samples in a template. \\
To properly evaluate our iconicity scores, a fair comparison with existing face quality metrics as well as a general IQA metric is required.  As mentioned in Section \ref{sec:prev}, FD score (i.e. scores assigned to a face by a face detector, effectively implying faceness) and norm of features are the only quality scores which do not require any reference image during training and evaluation. Therefore, we compare the quality scores of Model-1 and Model-2 with: a.) Face Detection scores \cite{ranjan2018crystal}, b.) Norm of features in \cite{sankaranarayanan2016triplet} and c.)BRISQUE \cite{mittal2012no} (an existing general IQA metric), for showing correlation with respect to affecting factors in most of the experiments. We cannot use Norm of \cite{ranjan2018crystal} as these features have unit norm. For the task of template based face verification, we compare the verification results obtained after quality pooling with scores of Model-1 and Model-2, with FD Score and Norm of \cite{sankaranarayanan2016triplet}. Finally we compare the performance breadth of each of these metrics.
\begin{figure}
\subfloat[]{\includegraphics[width = 0.5\linewidth, height = 30mm]{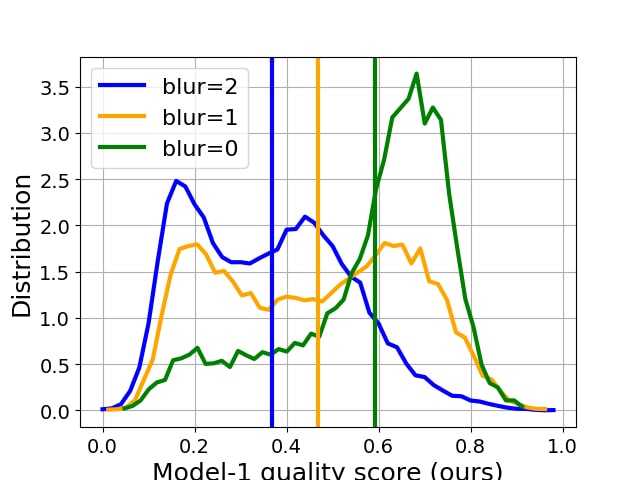}}
\subfloat[]{\includegraphics[width = 0.5\linewidth, height = 30mm]{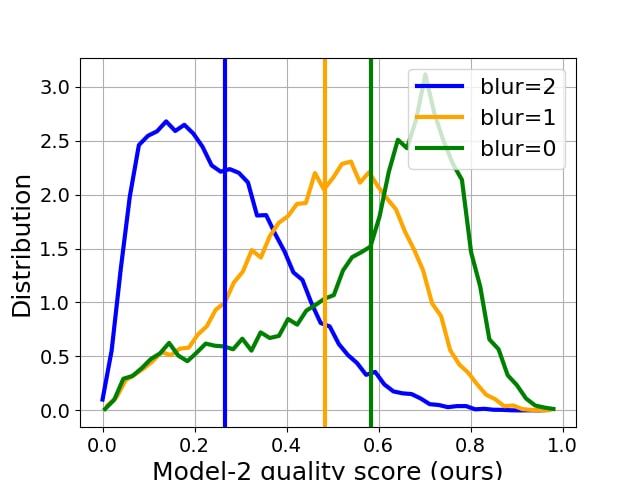}}\\ \vspace{-4mm}
\subfloat[]{\includegraphics[width = 0.5\linewidth, height = 30mm]{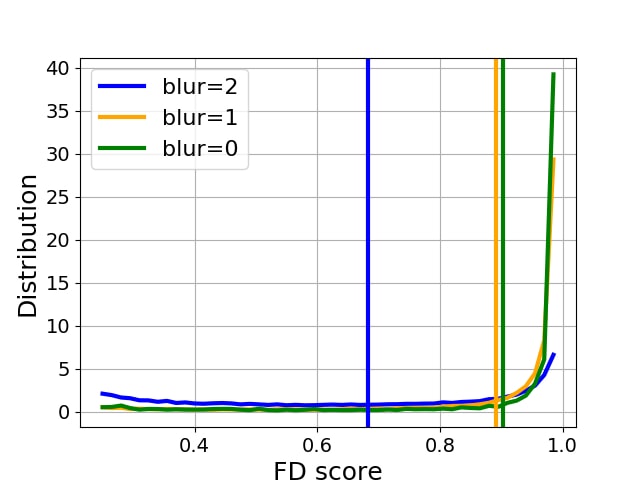}}
\subfloat[]{\includegraphics[width = 0.5\linewidth, height = 30mm]{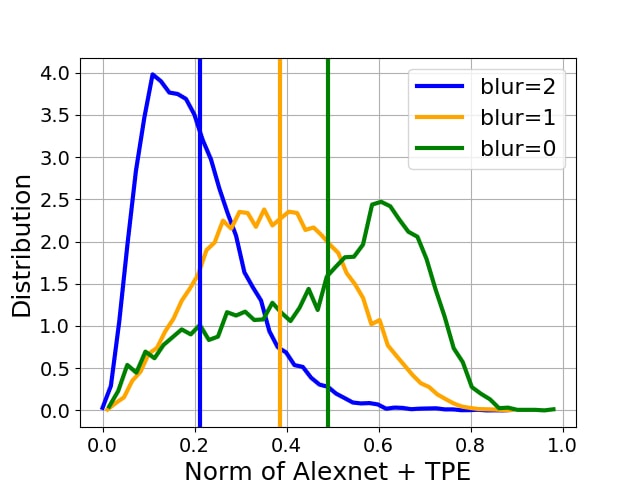}}\\ \vspace{-4mm}
\subfloat[]{\includegraphics[width = 0.5\linewidth, height = 30mm]{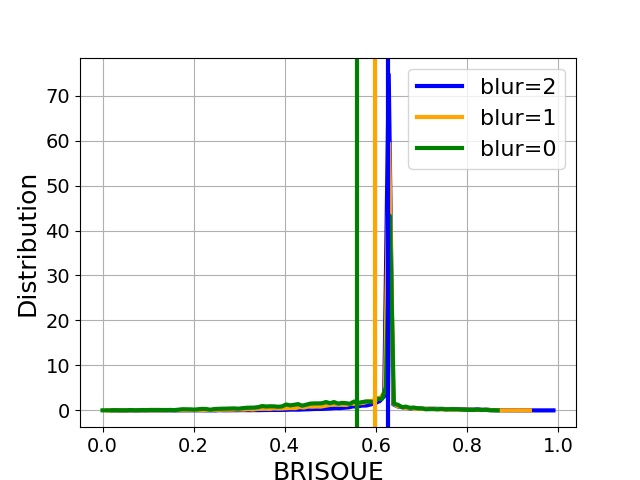}} 
\subfloat[]{\includegraphics[width = 0.5\linewidth, height = 30mm]{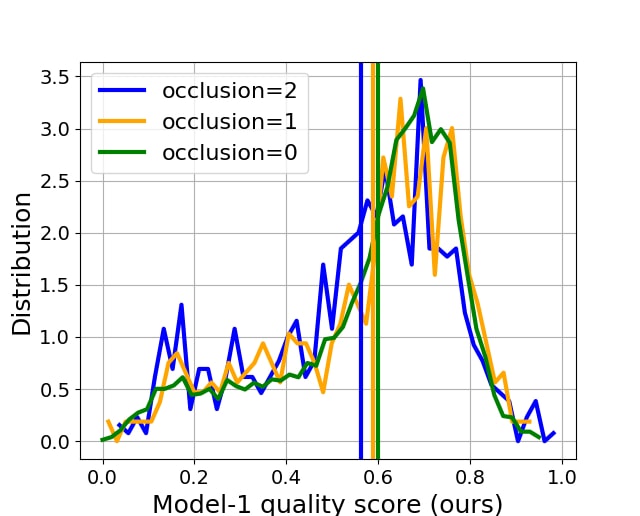}}\\ \vspace{-4mm}
\subfloat[]{\includegraphics[width = 0.5\linewidth, height = 30mm]{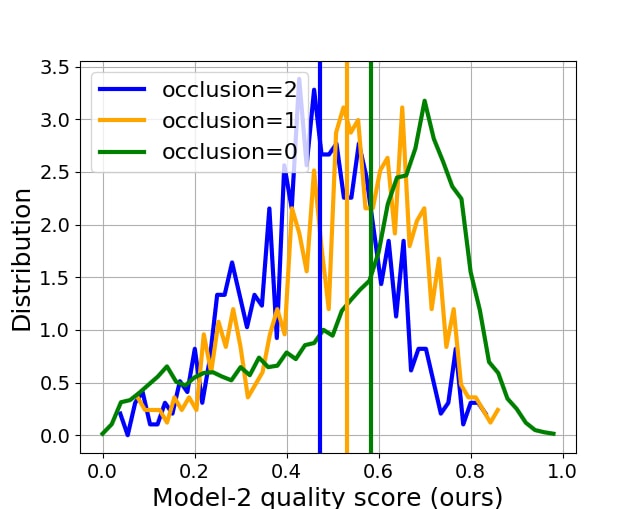}} 
\subfloat[]{\includegraphics[width = 0.5\linewidth, height = 30mm]{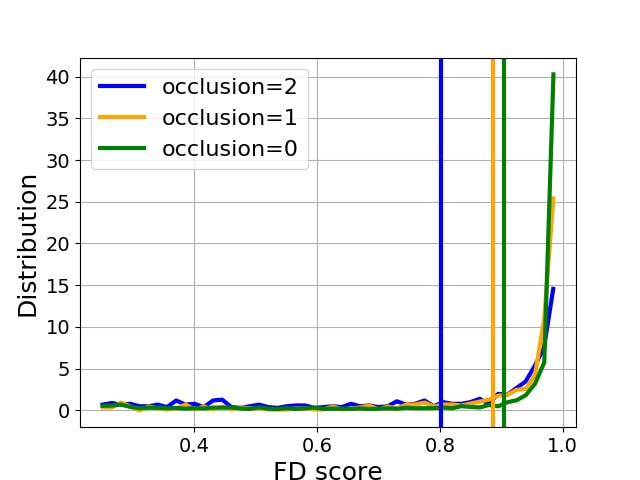}}\\
\subfloat[]{\includegraphics[width = 0.5\linewidth, height = 30mm]{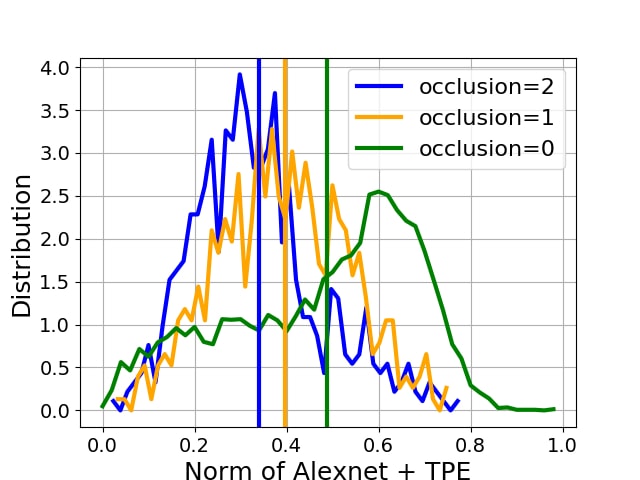}}
\subfloat[]{\includegraphics[width = 0.5\linewidth, height = 30mm]{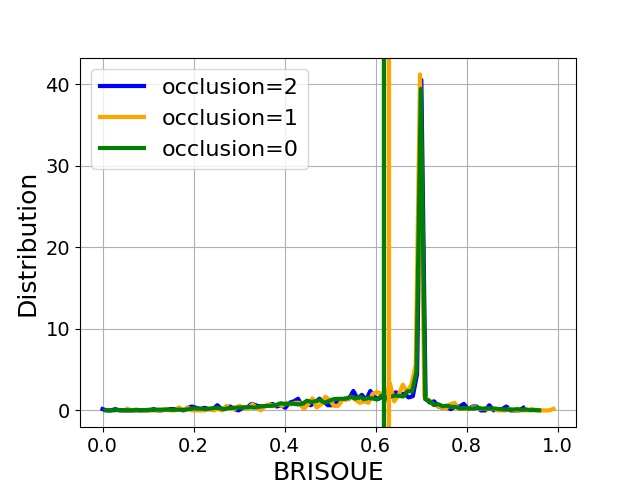}}\vspace{-4mm}
\caption{\textbf{(a-e)} Distributions of quality scores on different levels of blur obtained using a.) Model-1 b.) Model-2 c.) Face detection scores \cite{ranjan2018crystal}, d.) Norm of \cite{sankaranarayanan2016triplet} e.) BRISQUE. \textbf{(f-j)} Distributions of quality scores on different levels of occlusion obtained using f.) Model-1 g.) Model-2 h.) Face detection scores \cite{ranjan2018crystal} i.) Norm of \cite{sankaranarayanan2016triplet} j.) BRISQUE}
\label{fig:blurcorr}
\end{figure}
\vspace{-2mm}
\subsection{Visualization}
Firstly, we visualize the images according to Model-1 scores. Figure \ref{fig:visual} confirms that the scores accurately capture the visual quality of a facial image.
\subsection{Dependence on yaw, pitch and roll}
\vspace{-1.2mm}
The quality score of the facial image is expected to decrease as the pose of the face becomes extreme. We define pose on the basis of yaw, pitch and roll. For evaluation, we compute the values of these parameters for images in IJB-C dataset \cite{maze2018iarpa} using the all-in-one ConvNet trained in \cite{ranjan2017all}. The IJB-C dataset consists of 3531 identities with a total of 31,334 still images and 117,542 video frames collected in unconstrained settings. Following this we bin these values in a way such that each bin approximately contains equal number of images. For each bin, we compute the average quality score and visualize their variance as the pose keeps on becoming extreme. It is worth noting that while analyzing one pose parameters, the other two parameters are constrained to be between \ang{-30} to \ang{+30}. It can be confirmed from Figure \ref{fig:posecorr} that the Siamese MLP iconicity scores (Model-1 and Model-2) correlate well with pose variation. It should be emphasized that even though the model was not given any explicit information about facial poses, it is able to capture the relation of pose and quality on a dataset with a domain, completely different from that of the training dataset. On the other hand, BRISQUE \cite{mittal2012no}, being a general IQA metric, shows no correlation with face pose. Also from Figure \ref{fig:posecorr}, it is clear that the FD score does not correlate well with roll.
\vspace{-3.5mm}
\subsection{Dependence on blur}
The quality score of the facial image would decrease as the image gets degraded due to blur. To experimentally verify this using the Siamese MLP iconicity scores, we evaluate the scores on images in the WIDER Face dataset \cite{yang2016wider}. We use the training split of this dataset as we have ground truth annotation for blur of the image. The training images have been labeled to have one of the three blur labels : 0 - for no blur. 1 - for partial blur and 2 - for extreme blur. We plot the distribution of images belonging to each of these blur levels with respect to the score obtained from the Siamese MLP. We chose images with no extremes in terms of other parameters (such as occlusion, illumination, expression etc.). We test the score obtained by Model-1 and Model-2. It can be seen from  Figures \ref{fig:blurcorr}(a) and (b) that the average quality score (represented by the vertical line) keeps decreasing as the amount of blur is increased. In addition, the model is able to assign somewhat separate score distribution to different levels of blur. Here, the iconicity models were not provided with any explicit information about blur while being trained. We also provide corresponding plots using three other quality scores : Face Detection (FD) score \cite{ranjan2018crystal}, norm of the feature \cite{sankaranarayanan2016triplet} and BRISQUE \cite{mittal2012no}. We can infer that while norm of the feature models the blur quite well, the distributions obtained using BRISQUE and FD score are not resolvable. 
\vspace{-2mm}
 \subsection{Dependence on occlusion}
Facial image quality can be degraded due to possible occlusions. To corroborate the universality of the Siamese MLP scores, we observe the variation of the score with respect to occlusion and compare it with the score-variation of FD score and feature norm. We use the training split of WIDER dataset as we have ground truth annotation for occlusion of the image. Figure \ref{fig:blurcorr}(f-j) presents the distributions for different levels of occlusion modeled by Model-1, Model-2, Norm of \cite{sankaranarayanan2016triplet} and Face Detection scores \cite{ranjan2018crystal}. Once again it is clear from Figure \ref{fig:blurcorr}(h) and (j) that FD score and BRISQUE do not correlate well with occlusion. 
\vspace{-2mm}
\subsection{Template-based face verification}
\label{sec:fv}
\begin{figure}
    \includegraphics[width=\linewidth,height=35mm]{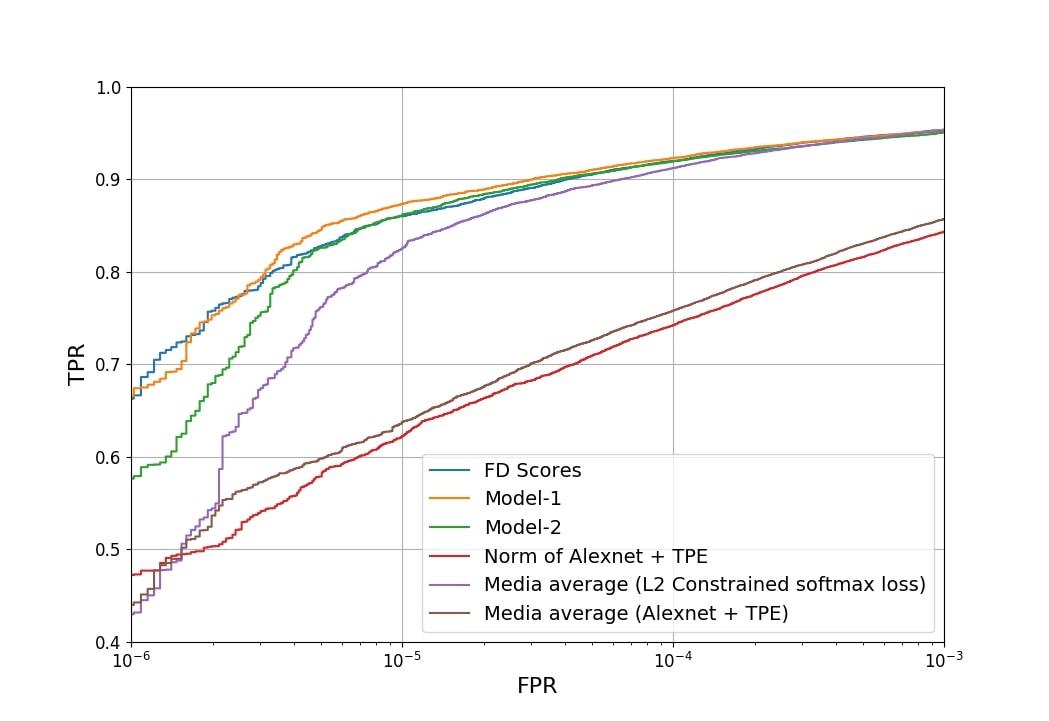}
    \caption{We perform face verification by pooling features with various quality scores, on IJB-C dataset}
    \label{fig:all_ver}
\end{figure}

\begin{table}
\centering
\small{
\begin{tabular}{l|l}
Features used & Scores for quality pooling \\ 
\hline

$L_2$ constrained softmax loss \cite{ranjan2017l2} & FD Score \\
$L_2$ constrained softmax loss \cite{ranjan2017l2}  & None (Media average)  \\
AlexNet \cite{sankaranarayanan2016triplet} & Feature Norm\\
AlexNet \cite{sankaranarayanan2016triplet} & None (Media average)\\
$L_2$ constrained softmax loss \cite{ranjan2017l2}  &Model-1\\
AlexNet \cite{sankaranarayanan2016triplet} &Model-2
\end{tabular}
\caption{Combination of features and quality pooling scores, used in this experiment} 
\label{table:combo}
}
\vspace{-3mm}
\end{table}
\begin{table}
\footnotesize{
\centering
\begin{tabular}{l|l|l|l|l|l|l}
Method / FPR: & $10^{-6}$ & $10^{-5}$ & $10^{-4}$ & $10^{-3}$ & $10^{-2}$ & $10^{-1}$ \\ 
\hline
Facenet \cite{schroff2015facenet} & 20.95& 33.30&48.69 &66.45 &81.76 &92.45 \\
VGGFace \cite{Parkhi15} &32.20 & 43.69&59.75 &74.79 &87.13 &95.64 \\
\hline
Norm of \cite{sankaranarayanan2016triplet} &47.2 & 61.9 &74.1 &84.2 &91.8 & 97.3\\
FD Score on \cite{ranjan2017l2} & 66.2& 85.9& 91.9& \textbf{95.3}& 97.5& 98.8\\
MediaAvg of \cite{ranjan2017l2} &43.7& 82.4& 91.1& 95.2& \textbf{97.6}& \textbf{98.9} \\
Model-1 (Ours) & \textbf{66.5}& \textbf{87.3}& \textbf{92.3}& \textbf{95.3}& 97.4& 98.7 \\
\end{tabular}
\caption{True positive rates for different face verification methods on the IJB-C data set.}\label{table:test}
}
\end{table}
After comparing the correlation of iconicity/quality metrics with various factors, we perform face verification by pooling features with various quality scores (including the scores of Model-1 and Model-2). The combination of features and corresponding pooling weights is mentioned in Table \ref{table:combo}. These experiments are performed on the IJB-C dataset, using the verification protocol specified in \cite{maze2018iarpa}. The verification protocol includes 19557 genuine matches and 15,638,932 impostor matches, which allows
us to evaluate the performance at very low FARs of $10^{-6}$. The algorithm used for quality pooling is same as in \cite{ranjan2018crystal}. Given feature $f_i$ in a template of $L$ images and corresponding quality score $r_i$ from a given model (Model-1 or Model-2), we compute $q_i = \frac{e^{\lambda r_i}}{\sum_{j=1}^L e^{\lambda r_j}}$ where $\lambda$ is a hyperparameter. We empirically chose $\lambda = 0.3$ in our experiments, based on the verification performance on the held-out data. Following this, we use $q_i$ to weight feature $f_i$ in a given template. We obtain the final feature descriptor as $f = \sum_{i=1}^L q_if_i$, and use it for verification. It is clear from Figure \ref{fig:all_ver} that Model-1 and Model-2 perform better than media averaging of features and norm of features in \cite{sankaranarayanan2016triplet}, especially at low FARs. Model-1's performance, is especially comparable to that of quality pooling features of \cite{ranjan2017l2} with FD scores. Moreover, it is observed that features of \cite{ranjan2017l2} outperform AlexNet features \cite{sankaranarayanan2016triplet} in general. Hence we pool features of \cite{ranjan2017l2} using scores of Model-1 (our best performing model), FD scores  and media averaging (which is our baseline) and compare their respective verification results in Table \ref{table:test}. Clearly, the iconicity scores from Model-1 outperform Norm of \cite{sankaranarayanan2016triplet} and media averaging of features of \cite{ranjan2017l2}, Facenet \cite{schroff2015facenet} and VGGFace \cite{Parkhi15}.
\vspace{-2mm}
\vspace{-1mm}

\subsection{Inference}
We now discuss the verification results obtained with our iconicity scores and analyze the difference between the results obtained with Model-1 and Model-2. Our approach to train iconicity models is dependent on the qualitative information of facial quality encoded in the feature. Quality-rich facial features would help to learn a better iconicity model, trained with such features. Hence we perform a small experiment to compare the quality information of faces present in the features of \cite{ranjan2017l2} and \cite{sankaranarayanan2016triplet}. In this experiment we compare the relative information of facial yaw in these features, since yaw is an important facial attribute that affects verifiability (and hence the iconicity) of a face.

We  randomly select 1000 features using \cite{ranjan2017l2} and \cite{sankaranarayanan2016triplet} and divide them into training and testing split (60\% and 40\% respectively). For this data, we also compute the yaws using \cite{ranjan2017all}. We then train a linear regression model using the training data, to predict facial yaw. Finally, we compare the relative error of the linear regressor trained with \cite{ranjan2017l2} and \cite{sankaranarayanan2016triplet} to estimate the amount of yaw information contained in these features. The regression errors are provided in Table \ref{table:err}. Clearly, the error of the regression model trained with \cite{sankaranarayanan2016triplet} is more than that trained with \cite{ranjan2017l2}. Hence we believe that features of \cite{ranjan2017l2} encapsulates much more yaw (and hence quality) information than \cite{sankaranarayanan2016triplet}. This explains the superiority of the iconicity model trained with \cite{ranjan2017l2} over that trained with \cite{sankaranarayanan2016triplet}.

\section{Performance breadth}
\begin{table}
\centering
\small{
\begin{tabular}{l|l}
Features used for linear regression & Test error \\ 
\hline

$L_2$ constrained softmax loss \cite{ranjan2017l2} & 0.71 \\
AlexNet \cite{sankaranarayanan2016triplet} & 0.84\\
\end{tabular}
\caption{Errors of linear regression models while predicting facial yaw} 
\label{table:err}
}
\vspace{-3mm}
\end{table}
For any ideal face quality metric, it is the performance breadth (rather than depth in certain tasks) and universality that demonstrates the efficiency of the metric.  We find that BRISQUE (which is a general IQA metric) is outperformed by all other methods. Citing the difference between IQA and face quality prediction, this outcome is expected. We find that FD scores can be effective to perform template based face verification, but it does not correlate well with blur, occlusion and roll (see Figures \ref{fig:posecorr} and \ref{fig:blurcorr} ). Also, we find that the norm of the features correlates well with yaw, pitch and other factors, but performs poorly when used for template based face verification, especially at low FARs (Table \ref{table:test}). Moreover, the norm cannot be used as a face quality metric if the features which have uniform norms. Hence, it is not universal. Interestingly, our iconicity scores (Model-1/Model-2) correlate well with all factors affecting the face quality and also obtains verification results comparable to state of the art. Thus, as the Siamese MLP scores demonstrate maximum performance breadth and is universal, it is closest to an ideal quality metric, among existing metrics. The breadth results are summarized in Table \ref{table:breadth}.
\begin{table}
\footnotesize{
\centering
\begin{tabular}{l|l|l|l|l|l|l|l}
Method : & Y & P & R & Blur& Occl\textsuperscript{n} & Verific\textsuperscript{n} & Universal \\ 
\hline
BRISQUE \cite{mittal2012no} & \xmark & \xmark & \xmark & \xmark & \xmark & - & \cmark \\
FD Score & \cmark & \cmark & \xmark & \xmark & \xmark & \cmark & \cmark \\
Norm of \cite{sankaranarayanan2016triplet}  & \cmark & \cmark & \cmark & \cmark & \cmark & \xmark & \xmark \\
Ours & \cmark & \cmark & \cmark & \cmark & \cmark & \cmark & \cmark \\
\end{tabular}
\caption{Performance breadth of different face quality metrics across various tasks : Correlation with Yaw (Y), Pitch(P), Roll(R), Blur, Occlusion, Quality Pooling for face verification,  Universality }\label{table:breadth}
}
\end{table}
\vspace{-1.5mm}

\section{Discussion}
In this work, we proposed a data driven approach to learn the iconicity of an image feature without the use of a predefined set of quality indicating images, or any external resource to aid the training of the model. As iconicity implies quality, we observe the variation of our model scores with respect to factors that affect the quality of an image. Finally, we use our scores to weight the features for template-based face verification. Our scores outperform the media averaging technique for the same and shows improvement over that achieved by scores obtained directly from a face detector. 
\section*{Acknowledgement}
This research is based upon work supported by the Office of the Director of National Intelligence (ODNI), Intelligence Advanced Research Projects Activity (IARPA), via IARPA R\&D Contract No. 2014-14071600012. The views and conclusions contained herein are those of the authors and should not be interpreted as necessarily representing the official policies or endorsements, either expressed or implied, of the ODNI, IARPA, or the U.S. Government. The U.S. Government is authorized to reproduce and distribute reprints for Governmental purposes notwithstanding any copyright annotation thereon.
{\small
\bibliographystyle{ieee}
\bibliography{egbib}
}

\end{document}